\documentclass[runningheads]{llncs}
\usepackage[T1]{fontenc}
\usepackage{graphicx}
\usepackage{subcaption}
\usepackage{todonotes}

\usepackage{pdflscape}

\begin{document}
%

\title{Methods for Generating Drift in Text Streams}
\titlerunning{Textual Drift Gen. Methods for Incr. Text Stream Classifiers' Perf. Eval.}

\titlerunning{Methods for Generating Drift in Text Streams}
%
\author{Cristiano M. Garcia\inst{1,3}\orcidID{0000-0002-7475-146X} \and
Alessandro L. Koerich\inst{2}\orcidID{0000-0001-5879-7014} \and
Alceu de S. Britto Jr.\inst{3,4}\orcidID{0000-0002-3064-3563} \and Jean Paul Barddal\inst{3}\orcidID{0000-0001-9928-854X}}
\authorrunning{C. M. Garcia et al.}
%
\institute{Instituto Federal de Santa Catarina (IFSC), Câmpus Caçador, Brazil \email{cristiano.garcia@ifsc.edu.br} \and
École de Technologie Supérieure (ÉTS), Université du Québec, Montréal, Canada\\
\email{alessandro.koerich@etsmtl.ca}\\\and
Pontifícia Universidade Católica do Paraná (PUCPR), Curitiba, Brazil\\
\email{\{alceu,jean.barddal\}@ppgia.pucpr.br} \and
Universidade Estadual de Ponta Grossa (UEPG), Ponta Grossa, Brazil
}
\maketitle              
\begin{abstract}
Systems and individuals produce data continuously. On the Internet, people share their knowledge, sentiments, and opinions, provide reviews about services and products, and so on. Automatically learning from these textual data can provide insights to organizations and institutions, thus preventing financial impacts, for example. To learn from textual data over time, the machine learning system must account for concept drift. Concept drift is a frequent phenomenon in real-world datasets and corresponds to changes in data distribution over time. For instance, a concept drift occurs when sentiments change or a word's meaning is adjusted over time. Although concept drift is frequent in real-world applications, benchmark datasets with labeled drifts are rare in the literature. To bridge this gap, this paper provides four textual drift generation methods to ease the production of datasets with labeled drifts. These methods were applied to Yelp and Airbnb datasets and tested using incremental classifiers respecting the stream mining paradigm to evaluate their ability to recover from the drifts. Results show that all methods have their performance degraded right after the drifts, and the incremental SVM is the fastest to run and recover the previous performance levels regarding accuracy and Macro F1-Score.



\keywords{Concept drift  \and Text drift generation \and Text stream mining}
\end{abstract}
\section{Introduction}

The Internet has become a rich environment in the last decades. Internet users can purchase products, chat with friends, and share photos, videos, and opinions, mostly through social media platforms. Authors, including Suprem and Pu \cite{suprem2019assed}, consider social media posts as social sensors. Thus, organizations, institutions, and governments can leverage the data available on the Internet to predict trends and understand users' behaviors, sentiments, and public opinions, to mention a few. Therefore, the scientific community's interest in developing methods for automatically learning from data available on the Internet has increased. More specifically, textual data produced on the Internet has been utilized for sentiment analysis \cite{bechini2021addressing}, crisis management \cite{pohl2018batch}, and fake reviews detection \cite{mohawesh2021analysis}.


Automatically learning from textual data has several challenges, such as standardization, vocabulary maintenance, etc. Additionally, the text vectorization method plays a crucial role, where good-quality representations can aid a machine learning model in reaching good performance levels. Text data analysis over time can bring additional difficulties, such as concept drift. Concept drift consists of data distribution changes over time \cite{gama2014survey}. Systems that leverage machine learning models without mechanisms to handle concept drift commonly have their performance levels degraded over time.


Several drift generation methods exist for traditional data streams \cite{breiman2017classification,ikonomovska2011learning}.
Simulations of concept drift for traditional data streams are easier to develop than for text drifts. The four drift categories described by Gama et al.~\cite{gama2014survey}, i.e., gradual, abrupt, incremental, and reoccurring, can be simulated by performing data distribution changes in arbitrary stream points. However, textual drifts are not straightforward since there are levels at which those changes can happen.
Textual drifts may occur at embedding, sentence, and word levels. Although embeddings are dense, numerical vectors, a simulation by changing arbitrary features, i.e., vector positions, may not make sense, especially for Word2vec~\cite{mikolov2013efficient} and BERT embeddings~\cite{devlin2019bert}, since it could create undefined meanings. Differently, changes in a sentence's class affect the previously learned relationship between input and output. Changes in words, e.g., replacing a particular word with its antonym, could change the sentence's sense.
In computational linguistics, researchers investigate semantic shifts, which are changes in word meaning \cite{bravo2022incremental,kutuzov2018diachronic}. However, semantic shifts are not in the scope of this paper since they concern a deeper level of linguistics. 

Therefore, mechanisms to artificially generate textual drifts can greatly help, making it possible to create benchmarks and ease the evaluation of novel text stream classifiers and concept drift detectors, for example. Textual datasets with labeled drifts are unusual in the literature \cite{garcia2023concept}. Authors generally mention the existence of drifts without determining the changing points. Differently, Garcia et al.~\cite{garcia2023event} leverage event-based posts to determine sentiment drifts. 

Thus, this paper describes four methods to generate concept drift in textual datasets to create a common benchmark for future developments in the area. These datasets could be used to evaluate novel text classification and concept drift detection methods, for example. From an original and voluminous dataset, several subsets can be generated considering the text drift generation methods.

In addition, this paper considers the stream mining paradigm, in which the complete dataset is not available at once; the texts are provided in the form of a stream, arrive one by one or in small batches, temporally ordered, and the text stream can be infinite \cite{bifet2023machine,gama2014survey}. This paradigm makes more sense since data is produced on the Internet at a high frequency. Therefore, receiving data over time is more reasonable than collecting all the available past data.


The contributions of this paper are three-fold: (a) a set of methods for generating drifts in textual datasets, (b) a comparison of incremental classifiers when handling drifted datasets, and (c) a new text dataset for machine learning experiments based on Inside Airbnb data, with a focus on concept drift. 

This paper is organized as follows: Section \ref{sec:background} presents a background with important concepts to understand the paper fully. Section \ref{sec:text-drift-gen-methods} describes the text drift generation methods proposed in this paper. Section \ref{sec:experimental-results} presents the results, diving into the experimental protocol, including the original datasets, parameters for the proposed text generation methods, the text vectorization method utilized, the incremental classifiers, their parameters and implementation, and the metrics. In addition, Section \ref{sec:results-discussion} shows the results and discussion, and finally, Section \ref{sec:conclusion} concludes this paper.

\section{Background}
\label{sec:background}
This section describes the text stream mining paradigm and concept drift phenomenon.

\subsection{Text Stream Mining}

Data streams, according to Bifet et al.~\cite{bifet2023machine}, are an abstraction that enables real-time analytics. Distinctly from batch scenarios, data streams have unique characteristics \cite{bifet2023machine,gama2014survey}: (a) their items arrive one by one or in small batches, meaning the data is not completely available at once; (b) their items arrive temporally ordered; and (c) the stream is potentially infinite.
To automatically learn from streaming environments, some abilities are desired from machine learning models, such as \cite{bifet2023machine,gama2014survey}: (a) the model must learn incrementally; (b) data must not be stored for a long time; ideally, the model should not store the data after learning from it; (c) the model must perform single-pass operations, and (d) the model must be conservative in the use of storage and time.

Text streams are a specialization of data streams \cite{garcia2023event}. Text streams are a potentially infinite sequence of texts, represented by $T = \{X_1, X_2, X_3, ...\}$, where $X$ is a text. In addition, a label $y$ may arrive associated with each $X$. X (former Twitter) posts, news, and service/product reviews are examples of text streams. 
Processing text streams frequently adds challenges related to natural language processing, such as text standardization and vocabulary maintenance. Similar to data streams, text streams are also subject to concept drift.

\subsection{Concept Drift}

The world is changing continuously, directly impacting data distribution. According to Gama et al.~\cite{gama2014survey}, concept drift corresponds to changes in data distribution over time. Formally, concept drift is generally described in the literature as $\exists$$x$ $:$ $p_{t} (y|X) \neq p_{t+\Delta} (y|x)$, where $\Delta > 0$, where $p_t$ is the conditional probability at a given time $t$. Gama et al.~\cite{gama2014survey} categorize types of concept drift into \textit{real concept drift} and \textit{virtual concept drift}, in which the former corresponds to changes in $p(y|X)$. Conversely, the latter refers to changes in $p(X)$ without necessarily affecting $y$. Considering the dynamics, Gama et al.~\cite{gama2014survey} consider four categories: (a) \textit{sudden/abrupt drift}, in which the concept drift happens with $\Delta = 1$; (b) \textit{incremental drift}, where $1 < \Delta < d$, where $d$ is an arbitrary number; (c) \textit{gradual changes}, in which the concepts switch to a new distribution and back, until it keeps in the new distribution; and (d) \textit{reoccurring changes}, in which the concepts switch distribution over time with variable $\Delta$.



Several drift generation methods for traditional data streams have been developed \cite{breiman2017classification,ikonomovska2011learning}. However, concept drifts in traditional data streams are simpler to simulate than in text streams. 
Changes in a traditional stream can be made by changing the data distribution of the process generator at a given position $t$, making it possible to simulate gradual, abrupt, incremental, or reoccurring drifts, following the aforementioned types described by Gama in \cite{gama2014survey}, for example. Concept drift generation methods for data streams can include drifts in given features, i.e., feature drift \cite{barddal2017survey}. 

In textual streams, the drifts have different characteristics compared to traditional domains. Textual drifts may be seen in embedding, sentences, and word levels. Regarding embeddings, dense vectors provided as representations by methods such as Word2Vec \cite{mikolov2013efficient} and BERT \cite{devlin2019bert}, changes in their positions could create undefined meanings. 
On the other hand, changes in a sentence's class affect the relationship between input, i.e., $X$, and output, i.e., $y$.
Changes in words, e.g., replacing a particular word with its antonym, could generate a (humanly-)noticeable change in the sentence's meaning.
A different type of change considers verifying different periods since the way of writing may change over time. Therefore, a textual drift can be generated by organizing texts from different time spans sequentially and temporally ordered. Given this contextualization, it can be concluded that most methods initially developed for traditional data streams cannot be applied to generate textual drift.

In computational linguistics, researchers investigate semantic shifts, which are changes in word meaning \cite{bravo2022incremental,kutuzov2018diachronic}. 
Referring to texts, \textit{semantic shift} corresponds to the ``evolution of a word meaning over time'' \cite{kutuzov2018diachronic}. Although semantic shift considers the word meaning, we bring forward this topic because the Adjective Swap method, presented in Section \ref{sec:text-drift-gen-methods} and utilized in this paper, could be perceived as a semantic shift since it reverses the original meaning of the sentence. Semantic shift studies generally involve the so-called \textit{diachronic} datasets, which contain texts from long time spans \cite{garcia2023concept,kutuzov2018diachronic}, focusing on specific words more than sentences. In addition, these studies rarely consider the streaming constraints \cite{garcia2023concept}, working mostly in batch processing. Therefore, since the semantic shift phenomenon is out of the scope, this paper uses the terminology \textit{concept drift} for the changes generated in the text drift generation methods.

\section{Text Drift Generation Methods}
\label{sec:text-drift-gen-methods}


Given the lack of text drift-generation methods and textual datasets with labeled drifts, this section presents four text drift-generating methods. We propose the class-based methods, while the Adjective Swap method was previously used in the literature to score a token sentiment drift \cite{bravo2022incremental}. They are detailed as follows.

\subsection{Class-based methods}
\label{subsec:class-methods}

We propose three class-based methods for generating text drifts in textual datasets: (a) Class Swap, (b) Class Shift, and (c) Time-slice Removal. In addition, we describe the Adjective Swap method, presented by Bravo-Marquez et al.~\cite{bravo2022incremental}, which changes the sense of a sentence.

\subsubsection{Class Swap}
Class Swap regards replacing instance classes after a certain period. For example, considering a 5-class textual dataset containing the classes $\{1, 2, 3, 4, 5\}$, the Class Swap method consists in swapping opposite classes, i.e., $1\leftrightarrow 5$ and $2 \leftrightarrow4$, thus becoming $\{5, 4, 3, 2, 1\}$, at an arbitrary time step $t$. This method aims at generating abrupt drifts. 
A limitation of the Class Swap method is that it makes more sense in multiclass scenarios. Fig.~\ref{fig:repr-class-swap} graphically demonstrates the Class Swap operations.

\subsubsection{Class Shift}
The Class Shift method considers a gradual change at arbitrary time steps $t$. Again, considering a hypothetical 5-class textual dataset containing the classes $\{1, 2, 3, 4, 5\}$, the Class Shift method changes a given class to the subsequent class, i.e., becoming $\{2, 3, 4, 5, 1\}$, at an arbitrary time step $t$.

\begin{figure*}[!htp]
    \centering
    \begin{subfigure}[t]{0.3\textwidth}
        \centering
        \includegraphics[width=\linewidth]{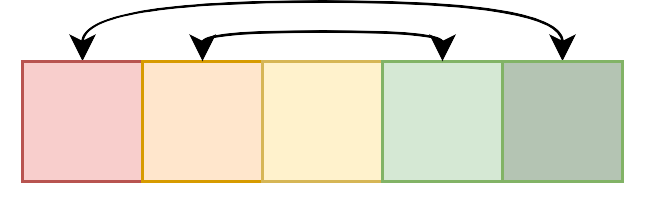}
        \caption{Class Swap method.}
        \label{fig:repr-class-swap}
    \end{subfigure}%
    ~ 
    \begin{subfigure}[t]{0.3\textwidth}
        \centering
        \includegraphics[width=\linewidth]{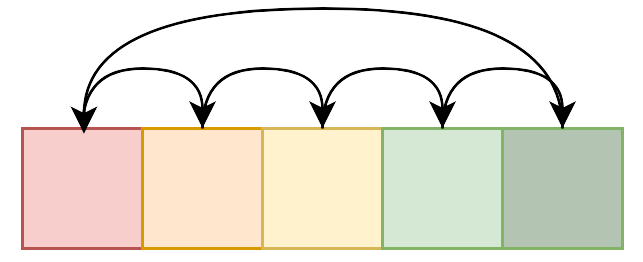}
        \caption{Class Shift method.}
        \label{fig:repr-class-shift}
    \end{subfigure}%
    \caption{Representations of the Class Swap and Class Shift methods. Each color represents a class.}
\end{figure*}

A clear limitation of Class Swap and Class Shift methods is that they function similarly if the original dataset is binary: switching the classes. Fig.~\ref{fig:repr-class-shift} demonstrates how the classes are switched over time. 

\subsubsection{Time-slice Removal}

Time-slice Removal is a method useful with datasets with timestamps. In this case, slices are deleted from the original dataset during the sampling. The rationale is to evaluate whether the changes in writing between two periods impact the classifier's performance.
In this paper, we selected three years at random to be removed from the sampling.


\begin{figure}[!htp]
    \centering
    \includegraphics[width=.5\linewidth]{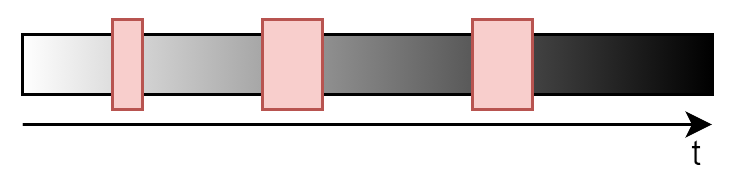}
    \caption{Representation of the Time Slice Removal method. The red squares represent randomly selected years to be removed from the text stream.}
    \label{fig:repr-tsr}
\end{figure}

\subsection{Adjective Swap}

Adjective Swap corresponds to changes in a sentence regarding sense or sentiment, for example. The main idea is to use a part-of-speech (POS) tagger to determine the role of the tokens, select the adjectives, and replace them with their antonyms when available. Using this method, the classes remain the same. In this paper, we leveraged a POS-tagger from the NLTK library \cite{bird2006nltk} and WordNet \cite{fellbaum1998wordnet} as the resource of word relations, tools frequently used in natural language processing tasks.

\begin{figure}[!htp]
    \centering
    \includegraphics[width=\linewidth]{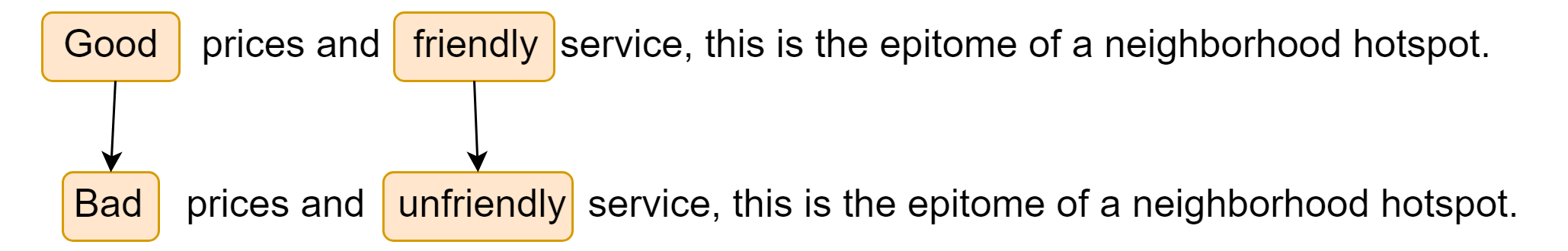}
    \caption{Representation of the Adjective Swap method. Orange rectangles represent the POS-tagged adjectives. The second row is the first one after applying the Adjective Swap method.}
    \label{fig:repr-adj-swap}
\end{figure}

This method aims to change the sentiment or sense of a sentence without replacing the class. Take the sentence ``Good prices and friendly service, this is the epitome of a neighborhood hotspot.'' as an example. The POS-tagger would return the words ``good'' and ``friendly'' as adjectives. Using WordNet to find potential antonyms for these words, it returns [`evil', `badness', `ill', `evilness', `bad'] and
[`unfriendly', `hostile'] as antonyms for ``good'' and ``friendly'', respectively. Performing the replacement, the resulting sentence may be ``Bad prices and unfriendly service, this is the epitome of a neighborhood hotspot.''. Fig. \ref{fig:repr-adj-swap} graphically demonstrates this example. Therefore, the sentences generated under Adjective Swap can bring difficulties to the incremental text stream classifier.

\section{Experimental Results}
\label{sec:experimental-results}


To evaluate the proposed text drift generation methods, we carried out several experiments. First, we have chosen a text dataset commonly used in machine learning experiments, i.e., the Yelp dataset, to apply the proposed drifts. In addition, we introduce a new text dataset based on the data available at the Inside Airbnb to evaluate the proposed drift methods as well. 


\subsection{Experimental Protocol}
\label{sec:experimental-protocol}
This section provides the experimental protocol, thus including the datasets, the dataset generation methods, the selected text vectorization method and incremental classifiers, and the metrics.

\subsection{Datasets}
This paper evaluates the proposed methods with 
the Yelp dataset, which contains textual information about businesses, reviews, and user data. Due to the lack of textual datasets for evaluating concept drift, we also introduce a new dataset derived from the Inside Airbnb, where we adapt part of the data for a classification task and further use it with the proposed drift generation methods.
These real-world datasets were chosen due to their sizes and the existence of timestamps/date fields to simulate text streams. In addition, Yelp is frequently used as a dataset for experiments in classification tasks \cite{mohawesh2021analysis}.
The Airbnb (New York) and Yelp datasets are described below.


\subsubsection{Airbnb}

The Airbnb dataset was collected from the Inside Airbnb website\footnote{http://insideairbnb.com/get-the-data/}. 
Inside Airbnb provides data sourced from publicly available information from the Airbnb site. The data from several countries has been analyzed, cleansed, and aggregated to facilitate public discussion. Unlike the Yelp dataset, the data is not readily available for any specific machine learning task. 

Therefore, we selected a subset related to New York. This choice happened because New York was the most visited city in 2022\footnote{Available at: https://www.cntraveler.com/story/most-visited-american-cities. Accessed on Jan 20th, 2024.} in the USA and, therefore, could provide a considerable volume of Airbnb reviews. We applied the following steps to adapt the data to a classification task, considering that the Airbnb dataset 
contains reviews written in several languages: (a) a pre-trained model for language identification is used (lid.176.ftz \cite{joulin2016bag,joulin2016fasttext}), and (b) a pre-trained model for sentiment analysis is applied to infer the reviews' sentiment (Twitter RoBERTa Base Sentiment\footnote{Available at: https://huggingface.co/cardiffnlp/twitter-roberta-base-sentiment.} model \cite{barbieri2020tweeteval}). After the reviews' languages are identified, the reviews are filtered to maintain only the reviews in English. Subsequently, the sentiments are estimated using the Twitter RoBERTa Base Sentiment model \cite{barbieri2020tweeteval}. 

After the treatment, 879,938 instances remained in the dataset. Fig. \ref{fig:airbnb-dataset-distr} shows the class distribution after filtering and keeping only reviews written in English. It is noticeable that this dataset is highly imbalanced, which brings interesting challenges to the compared incremental classifiers. We highlight that the proposed data preparation steps can be applied to any other subset of Inside Airbnb. We provide a script for this treatment on Github\footnote{https://github.com/cristianomg10/methods-for-generating-drift-in-text-streams}.



\subsubsection{Yelp}

Yelp Dataset\footnote{https://www.yelp.com/dataset} consists of a review collection regarding more than 150,000 businesses around the world located across 11 metropolitan areas. Besides the reviews, this dataset provides the number of stars corresponding to a given user's evaluation of a business, e.g., restaurants, hotels, etc. 

\begin{figure*}[t!]
    \centering
    \begin{subfigure}[t]{0.45\textwidth}
        \centering
        \includegraphics[width=\linewidth]{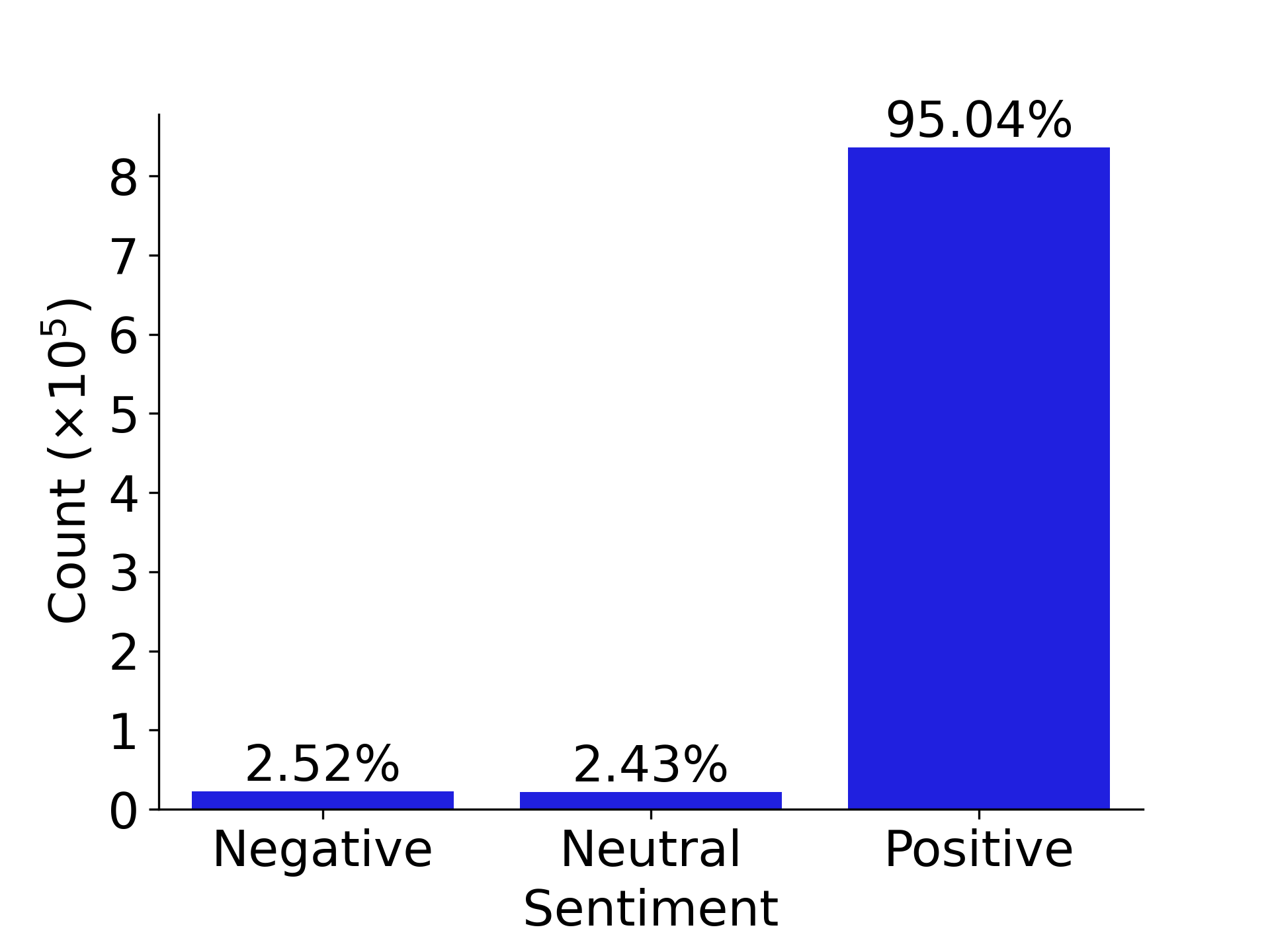}
        \caption{Class distribution of Airbnb dataset after filtering.}
        \label{fig:airbnb-dataset-distr}
    \end{subfigure}%
    ~ 
    \begin{subfigure}[t]{0.45\textwidth}
        \centering
        \includegraphics[width=\linewidth]{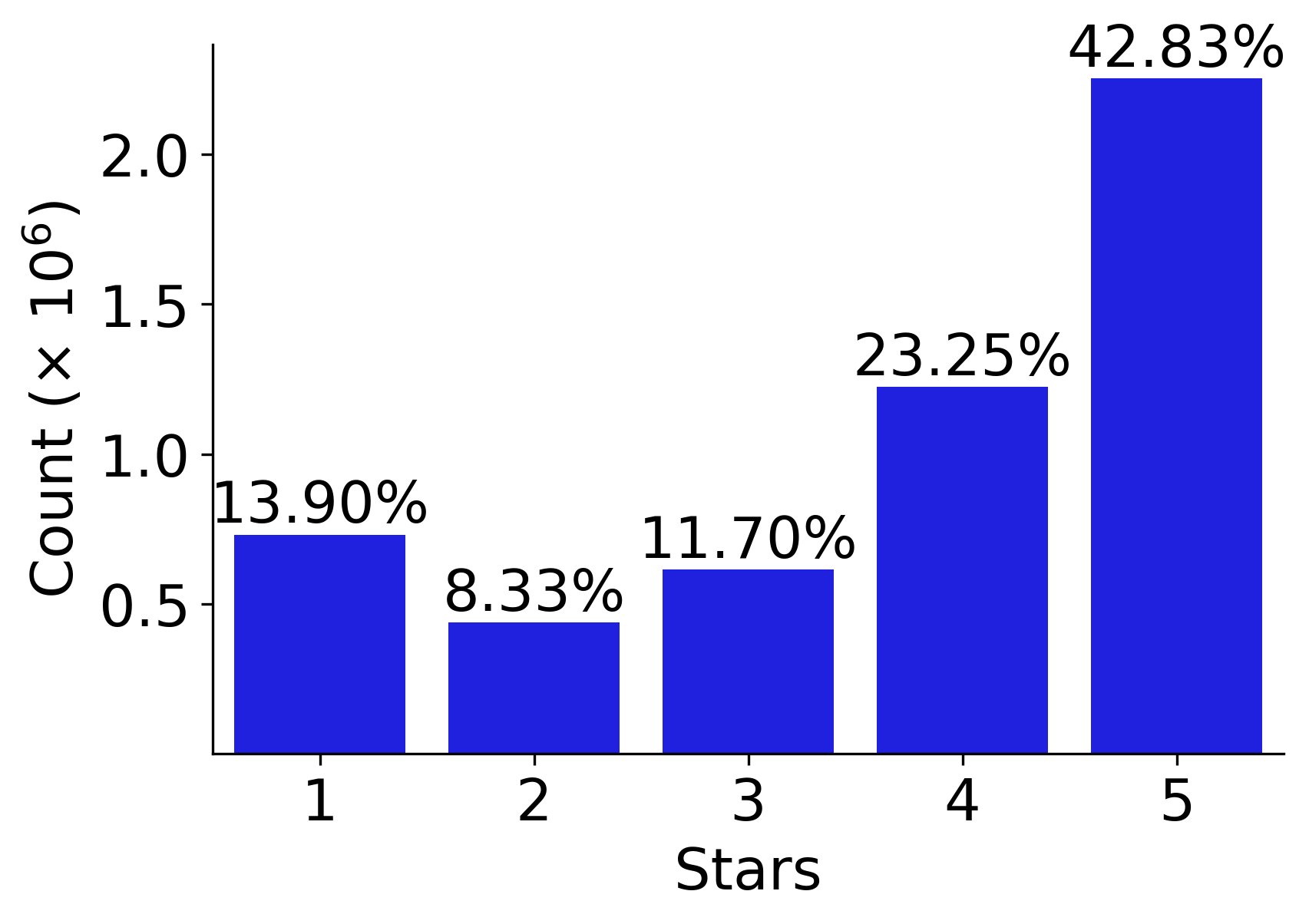}
        \caption{Class distribution of Yelp dataset. Adapted from \cite{thuma2023benchmarking}.}
        \label{fig:yelp-dataset-distr}
    \end{subfigure}%
    \caption{Class distribution of the datasets used in this paper.}
\end{figure*}

The version of the Yelp dataset used in this word contains 5,261,670 reviews distributed across five classes corresponding to stars. Fig. \ref{fig:yelp-dataset-distr} shows the class distribution for this dataset. This dataset also constitutes a challenge for the incremental classifiers since it is imbalanced, and slight variations, e.g., between 4 and 5 stars, are potentially difficult to capture.


\subsection{Subset Generation and Drift Generation Parameters}
Given the lack of textual datasets with labeled drifts, from the original Airbnb and Yelp datasets, ten subsets of length 200,000 are sampled for comparison purposes. To maintain the original distribution, the subsets are stratified and sampled by class, keeping the temporal order. The latter means respecting the stream characteristics in which data arrives in order \cite{bifet2023machine,gama2014survey}.

More specifically, the following steps are: (1) ten subsets are sampled from the original datasets; (2) each subset is copied four times, e.g., one copy per textual drift generation method, except for the Time-slice Removal method; and (3) the subset copies are applied to the text drift generation methods, except for the Time-slice Removal method, that has a separate sampling process. Therefore, since the same subsets originated in four out of five drifted subsets, a visual comparison is possible.

The parameters of the textual drift generation methods are as follows:

\begin{itemize}
    \item \textbf{Class Swap:} one swap at $t = 50000$;
    \item \textbf{Class Shift:} three shifts, at $t \in \{50000, 100000, 150000\}$;
    \item \textbf{Time-slice Removal: } three random years selected to be removed from the original sampling;
    \item \textbf{Adjective Swap: }one swap at $t = 50000$;
    \item \textbf{Adjective Swap: }three swaps, at $t \in \{50000, 100000, 150000\}$.
\end{itemize}

\subsection{Text Vectorization}

Text vectorization constitutes an operation that converts texts, considered non-structured data, into structured data \cite{thuma2023benchmarking}. This operation is essential since most machine learning methods expect structured data as input. 

In this paper, the selected text vectorization method is Sentence-BERT (SBERT) \cite{reimers2019sentencebert}, more specifically, the pre-trained model \texttt{paraphrase-MiniLM-L6-v2}. SBERT was initially developed using the BERT architecture to perform semantic textual similarity, although its provided embeddings could result in satisfactory performance in other tasks \cite{mokoatle2023review,thuma2023benchmarking}. This method encodes sentences into 384-number vectors, which can be used as input for the incremental text classification methods.

\subsection{Incremental Classifiers}

This paper evaluated three 
incremental classifiers: Gaussian Naive Bayes (GNB), Incremental Support Vector Machine (ISVM), and the Adaptive Random Forest (ARF) \cite{gomes2017adaptive}. 
Their distinct characteristics justify the choice for these classifiers: GNB considers data distributions, ISVM aims to find the best margins for data class separation, 
and ARF develops Hoeffding trees in an ensemble over time.
In our experiments, an ensemble of 10 trees was learned and updated.

\subsubsection{Implementation and Parameters}

Scikit-learn \cite{pedregosa2011scikit} and RiverML \cite{montiel2021river} libraries provided the implementations of the incremental classifiers used in this paper.
The selected parameters are determined as follows:

\begin{itemize}
    \item \textbf{GNB:} No parameters needed (RiverML);
    \item \textbf{ISVM:} Using the Stochastic Gradient Descent from Scikit-learn, with default parameters, except for the loss function, set as \texttt{hinge}, thus providing a linear SVM;
    \item \textbf{ARF:} ARF has an internal mechanism for ensemble update in the presence of concept drift. We tested ARF both with and without warning and drift detectors. The model's label is \textbf{ARF (DD)} whenever these mechanisms are enabled. ADWIN \cite{bifet2007learning} is used for drift detection in ARF (DD). By default, RiverML's implementation uses 10 Hoeffdings trees. 
    
\end{itemize}

\subsection{Hardware}
The hardware used in the experiments is a 13th-generation Intel(R) Core(TM) i9-13900K with 128 GB of RAM running Ubuntu 22.04 LTS.

\subsection{Metrics}
In this paper, we selected regular imbalanced multiclass classification metrics: (a) Accuracy, (b) Macro F1, and (b) elapsed time.

\begin{itemize}
    \item \textbf{Accuracy:} calculates the proportion between correct classifications and the total number of classified items;
    \item \textbf{Macro F1:} this metric calculates the F1 score per class and weights them equally; thus, the majority class is not favored by the metric, becoming thus suitable for imbalanced scenarios;
    \item \textbf{Elapsed time:} measured in seconds. 
\end{itemize}

Considering the stream paradigm, the windowed Macro F1 
will also be displayed over time.



\subsection{Results and Discussion}
\label{sec:results-discussion}

\begin{landscape}

\begin{figure}[!htp]
\centering
\includegraphics[width=\linewidth]{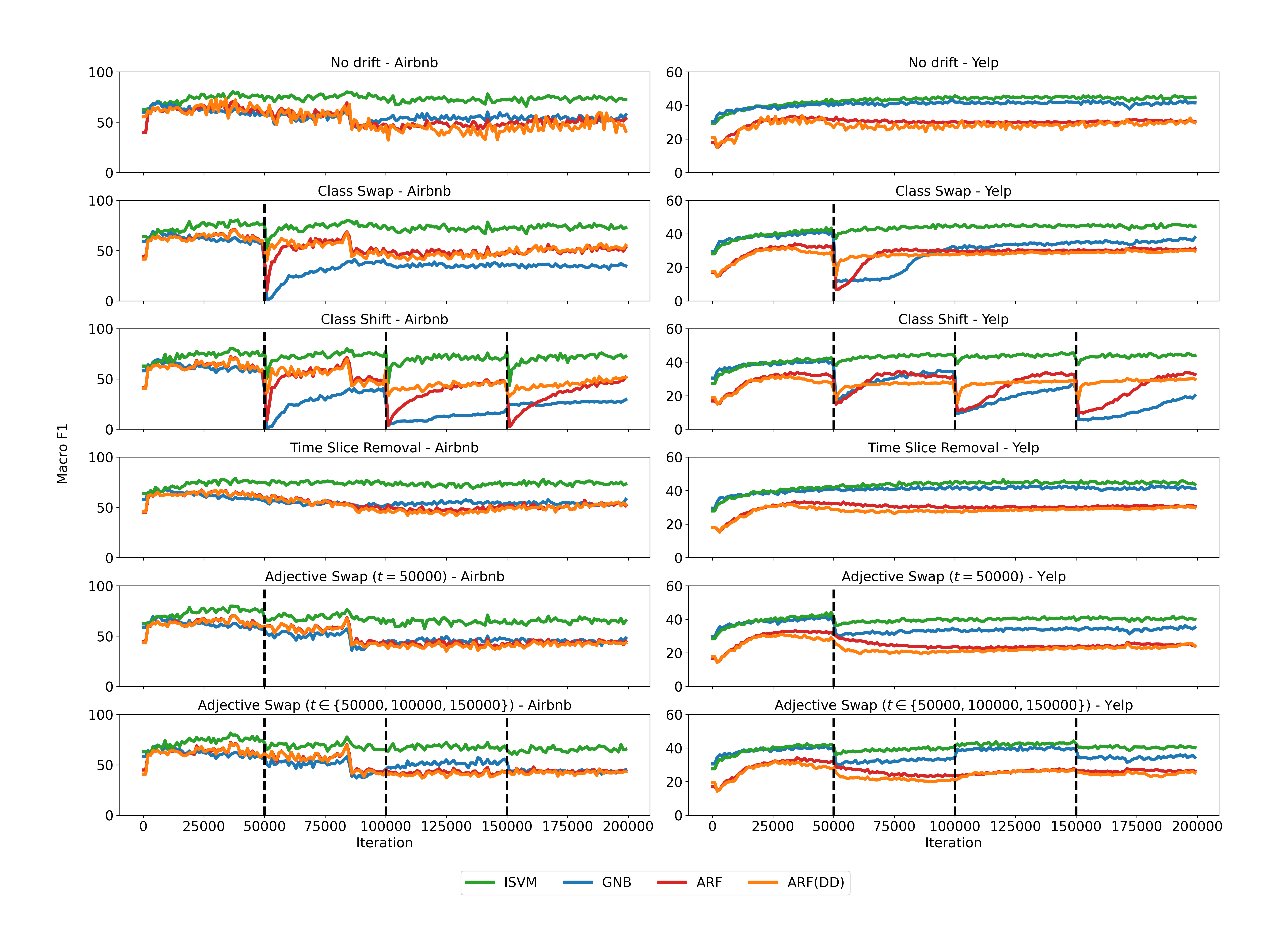}
\caption{Results for Airbnb (left) and Yelp (right). The dashed lines correspond to the drifts.}
\label{fig:results}
\end{figure}

\end{landscape}

First, we evaluated the Macro F1 score over time to visually assess the impact of the drifts generated by the methods. Fig.~\ref{fig:results} (best viewed in color) shows the Macro F1 score over time for the Airbnb and Yelp datasets on the left and right, respectively. 
The metric is computed and reset at every 1000 instances to perceive the short-term variations better. 
We also included a scenario with no artificially generated drift for reference.

Fig.~\ref{fig:results} (left) shows that all classifiers are impacted in the Class Swap and Class Shift scenarios in the Airbnb dataset. GNB is the most impacted, and ARF recovers faster than GNB from the drifts. ISVM is barely affected by the drifts, being also the classifier that recovers the fastest. This happens due to the fast adaptation of its support vectors. After the generated drifts, the performance of GNB increases slowly. ARF and ARF(DD) are also impacted. However, as expected, ARF(DD) recovers faster than ARF. In the Time Slice Removal, the classifiers perform steadily, while for the Adjective Swap (with three swaps), the first and the third drifts generated more impact. In addition, in the Adjective Swap scenario, the classifiers followed the same pattern considering the period after the single drift, compared to the Adjective Swaps (with three swaps). It is important to notice a ``natural'' drift around $t=80000$, and its impact is visible in all scenarios except in the Time-slice Removal scenario. That occurrence reinforces that drifts are frequent in textual datasets and highlights the importance of text drift generation methods for developing the text stream mining area.

Fig. \ref{fig:results} (right) shows the results for the Yelp dataset. GNB, ARF, and ARF(DD) are similarly impacted by the drifts in the Class Swap scenario, while ISVM had smaller variations right after the drift. As expected, ARF(DD) recovers faster than ARF and GNB. In the Class Shift scenario, ISVM had decreases in performance after the drifts but quickly recovered from them. On the other hand, GNB and ARF are highly impacted by the drifts, and ARF recovers faster than GNB, being ARF(DD) faster than both. In the Time-Slice Removal scenario, all the classifiers perform steadily. For the Adjective Swap, a decrease in the performance is observable in all classifiers after drift. Finally, the Adjective Swap (with three swaps) scenario shows an interesting behavior: a decrease is noticeable after the first and third drifts, while an increase in the performance happens after the second drift. We hypothesize that the classifiers, especially ISVM and GNB, have not completely forgotten the previous concepts.



In addition to the results over time, we consolidate the performance obtained regarding accuracy, Macro F1, and Elapsed time in Table \ref{tab:results}. The bold values are the best considering a drift scenario and a dataset. To better visualize the information, we reduced the names of the scenarios, e.g., Class Swap became C. Sw. and Adjective Swap with three swaps became Adj. Sw. (3). For all the scenarios, ISVM demonstrated to be very efficient, achieving the best results among the classifiers not only in terms of accuracy and Macro F1 but also regarding elapsed time. The metrics were calculated incrementally.

\begin{table}[!htp]
\centering
\caption{Performance of incremental classifiers grouped by dataset and scenarios. Bold values are the best for a given pair dataset scenario.}
\label{tab:results}
\resizebox{\linewidth}{!}{
\begin{tabular}{llllll}
\hline
\textbf{Method} & \textbf{Dataset} & \textbf{Scenario} & \textbf{Accuracy}  & \textbf{Macro F1} & \textbf{Elapsed time} \\ \hline
ARF 		 & Airbnb	 & No drift	 & 96.27 $\pm$ 0.03 & 58.77 $\pm$ 0.50 	& 835.56 $\pm$ 7.87 \\
ARF(DD) 	 & Airbnb	 & No drift	 & 96.22 $\pm$ 0.05 & 56.65 $\pm$ 0.00 	& 941.27 $\pm$ 27.70 \\
GNB 		 & Airbnb	 & No drift	 & 84.65 $\pm$ 0.17 & 56.33 $\pm$ 0.16 	& 599.32 $\pm$ 3.63 \\
ISVM 		 & Airbnb	 & No drift	 & \scalebox{0.94}{\textbf{97.24 $\pm$ 0.03}} & \scalebox{0.94}{\textbf{75.12 $\pm$ 0.25}} 	& \scalebox{0.94}{\textbf{505.46 $\pm$ 15.17}} \\\hline
ARF 		 & Airbnb	 & C. Sw.	 & 93.98 $\pm$ 0.09 & 71.14 $\pm$ 0.70 	& 849.71 $\pm$ 7.25 \\
ARF(DD) 	 & Airbnb	 & C. Sw.	 & 96.17 $\pm$ 0.02 & 73.26 $\pm$ 0.43 	& 985.32 $\pm$ 38.36 \\
GNB 		 & Airbnb	 & C. Sw.	 & 53.85 $\pm$ 0.15 & 46.34 $\pm$ 0.22 	& 574.41 $\pm$ 1.00 \\
ISVM 		 & Airbnb	 & C. Sw.	 & \scalebox{0.94}{\textbf{97.20 $\pm$ 0.03}} & \scalebox{0.94}{\textbf{86.77 $\pm$ 0.09}} 	& \scalebox{0.94}{\textbf{494.49 $\pm$ 4.45}} \\\hline
ARF 		 & Airbnb	 & C. Sh.	 & 82.57 $\pm$ 0.18 & 81.26 $\pm$ 0.20 	& 905.09 $\pm$ 6.11 \\
ARF(DD) 	 & Airbnb	 & C. Sh.	 & 95.90 $\pm$ 0.03 & 95.58 $\pm$ 0.02 	& 972.82 $\pm$ 30.42 \\
GNB 		 & Airbnb	 & C. Sh.	 & 48.99 $\pm$ 0.33 & 41.56 $\pm$ 0.41 	& 574.24 $\pm$ 1.18 \\
ISVM 		 & Airbnb	 & C. Sh.	 & \scalebox{0.94}{\textbf{97.06 $\pm$ 0.03}} & \scalebox{0.94}{\textbf{96.82 $\pm$ 0.03}} 	& \scalebox{0.94}{\textbf{494.82 $\pm$ 3.72}} \\\hline
ARF 		 & Airbnb	 & T. S. Rem.	 & 96.26 $\pm$ 0.09 & 58.76 $\pm$ 0.90 	& 827.97 $\pm$ 4.88 \\
ARF(DD) 	 & Airbnb	 & T. S. Rem.	 & 96.20 $\pm$ 0.11 & 57.55 $\pm$ 1.05 	& 1003.27 $\pm$ 43.79 \\
GNB 		 & Airbnb	 & T. S. Rem.	 & 84.44 $\pm$ 0.50 & 56.04 $\pm$ 0.88 	& 574.22 $\pm$ 1.66 \\
ISVM 		 & Airbnb	 & T. S. Rem.	 & \scalebox{0.94}{\textbf{97.22 $\pm$ 0.07}} & \scalebox{0.94}{\textbf{75.17 $\pm$ 0.38}} 	& \scalebox{0.94}{\textbf{494.07 $\pm$ 4.17}} \\\hline
ARF 		 & Airbnb	 & Adj. Sw.	 & 96.14 $\pm$ 0.02 & 56.47 $\pm$ 0.33 	& 837.17 $\pm$ 2.62 \\
ARF(DD) 	 & Airbnb	 & Adj. Sw.	 & 96.07 $\pm$ 0.02 & 55.12 $\pm$ 0.29 	& 1038.42 $\pm$ 38.57 \\
GNB 		 & Airbnb	 & Adj. Sw.	 & 79.13 $\pm$ 0.18 & 49.70 $\pm$ 0.33 	& 579.17 $\pm$ 0.96 \\
ISVM 		 & Airbnb	 & Adj. Sw.	 & \scalebox{0.94}{\textbf{96.62 $\pm$ 0.02}} & \scalebox{0.94}{\textbf{69.86 $\pm$ 0.16}} 	& \scalebox{0.94}{\textbf{493.11 $\pm$ 0.84}} \\\hline
ARF 		 & Airbnb	 & Adj. Sw. (3)	 & 96.14 $\pm$ 0.02 & 56.56 $\pm$ 0.32 	& 834.07 $\pm$ 6.44 \\
ARF(DD) 	 & Airbnb	 & Adj. Sw. (3)	 & 96.09 $\pm$ 0.02 & 55.48 $\pm$ 0.29 	& 1040.67 $\pm$ 51.86 \\
GNB 		 & Airbnb	 & Adj. Sw. (3)	 & 79.96 $\pm$ 0.27 & 50.01 $\pm$ 0.14 	& 579.33 $\pm$ 0.95 \\
ISVM 		 & Airbnb	 & Adj. Sw. (3)	 & \scalebox{0.94}{\textbf{96.70 $\pm$ 0.02}} & \scalebox{0.94}{\textbf{70.67 $\pm$ 0.25}} 	& \scalebox{0.94}{\textbf{492.61 $\pm$ 0.72}} \\\hline
ARF 		 & Yelp	 & No drift	 & 56.31 $\pm$ 0.22 & 28.08 $\pm$ 0.17 	& 1312.56 $\pm$ 33.00 \\
ARF(DD) 	 & Yelp	 & No drift	 & 55.35 $\pm$ 0.12 & 26.62 $\pm$ 0.00 	& 1465.83 $\pm$ 41.02 \\
GNB 		 & Yelp	 & No drift	 & 46.46 $\pm$ 0.16 & 34.42 $\pm$ 0.08 	& 722.88 $\pm$ 1.79 \\
ISVM 		 & Yelp	 & No drift	 & \scalebox{0.94}{\textbf{58.22 $\pm$ 0.11}} & \scalebox{0.94}{\textbf{44.58 $\pm$ 0.11}} 	& \scalebox{0.94}{\textbf{560.39 $\pm$ 3.20}} \\\hline
ARF 		 & Yelp	 & C. Sw.	 & 53.53 $\pm$ 0.18 & 28.08 $\pm$ 0.09 	& 1213.69 $\pm$ 17.06 \\
ARF(DD) 	 & Yelp	 & C. Sw.	 & 55.14 $\pm$ 0.16 & 27.66 $\pm$ 0.13 	& 1579.59 $\pm$ 62.97 \\
GNB 		 & Yelp	 & C. Sw.	 & 38.44 $\pm$ 0.20 & 28.41 $\pm$ 0.18 	& 695.51 $\pm$ 4.84 \\
ISVM 		 & Yelp	 & C. Sw.	 & \scalebox{0.94}{\textbf{58.05 $\pm$ 0.11}} & \scalebox{0.94}{\textbf{45.81 $\pm$ 0.11}} 	& \scalebox{0.94}{\textbf{572.06 $\pm$ 16.11}} \\\hline
ARF 		 & Yelp	 & C. Sh.	 & 41.55 $\pm$ 0.23 & 34.28 $\pm$ 0.19 	& 1243.35 $\pm$ 12.68 \\
ARF(DD) 	 & Yelp	 & C. Sh.	 & 54.45 $\pm$ 0.23 & 43.91 $\pm$ 0.20 	& 1590.93 $\pm$ 105.27 \\
GNB 		 & Yelp	 & C. Sh.	 & 27.14 $\pm$ 0.25 & 22.64 $\pm$ 0.20 	& 695.48 $\pm$ 3.28 \\
ISVM 		 & Yelp	 & C. Sh.	 & \scalebox{0.94}{\textbf{57.62 $\pm$ 0.12}} & \scalebox{0.94}{\textbf{56.36 $\pm$ 0.13}} 	& \scalebox{0.94}{\textbf{572.57 $\pm$ 12.79}} \\\hline
ARF 		 & Yelp	 & T. S. Rem.	 & 56.06 $\pm$ 0.94 & 28.31 $\pm$ 0.46 	& 1211.13 $\pm$ 10.24 \\
ARF(DD) 	 & Yelp	 & T. S. Rem.	 & 55.18 $\pm$ 1.08 & 26.91 $\pm$ 0.58 	& 1571.73 $\pm$ 77.59 \\
GNB 		 & Yelp	 & T. S. Rem.	 & 46.41 $\pm$ 0.48 & 34.43 $\pm$ 0.16 	& 697.14 $\pm$ 2.53 \\
ISVM 		 & Yelp	 & T. S. Rem.	 & \scalebox{0.94}{\textbf{58.01 $\pm$ 0.93}} & \scalebox{0.94}{\textbf{44.63 $\pm$ 0.19}} 	& \scalebox{0.94}{\textbf{573.39 $\pm$ 10.23}} \\\hline
ARF 		 & Yelp	 & Adj. Sw.	 & 52.06 $\pm$ 0.11 & 24.10 $\pm$ 0.14 	& 1194.48 $\pm$ 14.22 \\
ARF(DD) 	 & Yelp	 & Adj. Sw.	 & 51.13 $\pm$ 0.21 & 22.30 $\pm$ 0.26 	& 1605.34 $\pm$ 112.08 \\
GNB 		 & Yelp	 & Adj. Sw.	 & 39.80 $\pm$ 0.16 & 29.44 $\pm$ 0.07 	& 700.11 $\pm$ 3.79 \\
ISVM 		 & Yelp	 & Adj. Sw.	 & \scalebox{0.94}{\textbf{55.00 $\pm$ 0.16}} & \scalebox{0.94}{\textbf{41.26 $\pm$ 0.16}} 	& \scalebox{0.94}{\textbf{553.75 $\pm$ 2.57}} \\\hline
ARF 		 & Yelp	 & Adj. Sw. (3)	 & 52.96 $\pm$ 0.20 & 25.12 $\pm$ 0.18 	& 1196.83 $\pm$ 9.21 \\
ARF(DD) 	 & Yelp	 & Adj. Sw. (3)	 & 52.16 $\pm$ 0.24 & 23.69 $\pm$ 0.17 	& 1599.55 $\pm$ 60.29 \\
GNB 		 & Yelp	 & Adj. Sw. (3)	 & 41.48 $\pm$ 0.13 & 30.65 $\pm$ 0.13 	& 701.76 $\pm$ 2.14 \\
ISVM 		 & Yelp	 & Adj. Sw. (3)	 & \scalebox{0.94}{\textbf{55.56 $\pm$ 0.11}} & \scalebox{0.94}{\textbf{41.81 $\pm$ 0.13}} 	& \scalebox{0.94}{\textbf{555.04 $\pm$ 0.53}} \\\hline
\end{tabular}
}
\end{table}

In addition, we see that ISVM performed the task between 40 and 60\% of the time required by ARF. GNB performs faster than ARF; however, its performance levels are much lower than ARF's and SVM's. As expected, ARF(DD) is slightly slower than ARF, due to the ensemble adaptation mechanism.

\section{Conclusion}
\label{sec:conclusion}


Processing textual datasets, mainly as a stream, is nowadays necessary due to the real-time nature of several domains, such as social media and news. Although several authors, e.g., Heusinger et al. \cite{heusinger2020analyzing} mention the natural and frequent presence of different types of drift in textual streams, the available datasets with labeled drifts are rare.

This paper presents four methods for textual drift generation. Due to the lack of textual datasets with labeled drifts, those methods can help develop the text stream mining area. In addition, we present the Airbnb dataset obtained from Inside Airbnb. The dataset was processed, and reviews had their languages identified with a pre-trained model. The reviews' sentiments were also analyzed with a pre-trained RoBERTa Base Sentiment model.
The generated subsets can be used as a benchmark for future developments in the area, such as novel incremental classifiers, drift detectors, and so on. 

In this paper, Incremental SVM (ISVM), Gaussian Naive Bayes (GNB), and Adaptive Random Forest (ARF), with and without the drift adaptation mechanism enabled, were evaluated in five scenarios under drift. ISVM performed the best in all scenarios, suffering little impact when receiving drifted streams. On the other hand, ARF and GNB had more difficulty than ISVM, although ARF showed a better recovery ability than GNB. Furthermore, ARF(DD) could recover from the drifts faster than ARF and GNB.

As limitations, some methods may seem to impose unrealistic scenarios. However, the intention is to provide and make it possible to create a benchmark with different, challenging, and drifting scenarios, keeping the temporal order (an important characteristic of text streams). In addition, as observed in the experiment related to Airbnb and well-known in the literature \cite{heusinger2020analyzing}, drifts naturally appear in textual data streams.

In future works, we intend to evaluate different machine learning models in tasks such as text stream clustering under drifting text streams, considering also using drift detectors. Also, we intend to prepare novel drifted datasets from different domains for text stream applications.

\section*{Acknowledgements}

Cristiano Mesquita Garcia is a grantee of a \textit{Doutorado Sanduíche} scholarship provided by \textit{Fundação Coordenação de Aperfeiçoamento de Pessoal de Nível Superior} (CAPES).

%
%
%
\bibliographystyle{splncs04}
\bibliography{reference}
%




\end{document}